\documentclass[journal,onecolumn]{IEEEtran}
\makeatletter
\def\ps@headings{\def\@oddhead{\IEEEdoarxivheader{-1\oddsidemargin}\relax
\hbox{}\@IEEEheaderstyle\rightmark\hfil\thepage}\relax
\def\@evenhead{\IEEEdoarxivheader{-1\evensidemargin}\relax
\hbox{}\@IEEEheaderstyle\rightmark\hfil\thepage}\relax
\def\@oddfoot{\IEEEdoarxivfooter{-1\oddsidemargin}\hfil\hbox{}}\relax
\def\@evenfoot{\IEEEdoarxivfooter{-1\evensidemargin}\hfil\hbox{}}\relax}
\def\ps@IEEEtitlepagestyle{\ps@headings}

\def\IEEEarxivheadfootoffset{3pt}
\newdimen\IEEEheadtotopofpage
\newdimen\IEEEfoottobottomofpage
\newbox\@IEEEboxX

\def\IEEEarxivheader{}
\def\IEEEarxivfooter{}

\advance\IEEEheadtotopofpage by 1\topmargin
\advance\IEEEheadtotopofpage by 1in\relax

\IEEEfoottobottomofpage\paperheight
\advance\IEEEfoottobottomofpage by -1in\relax
\advance\IEEEfoottobottomofpage by -1\topmargin
\advance\IEEEfoottobottomofpage by -1\headheight
\advance\IEEEfoottobottomofpage by -1\headsep
\advance\IEEEfoottobottomofpage by -1\textheight
\advance\IEEEfoottobottomofpage by -1\footskip
\def\IEEEarxivheaderstyle{\normalfont\footnotesize}

\def\IEEEdoarxivheader#1{\@IEEEtrantmpdimenA\IEEEarxivheadfootoffset\relax
\@IEEEtrantmpdimenA -1\@IEEEtrantmpdimenA
\advance\@IEEEtrantmpdimenA by \IEEEheadtotopofpage
\settoheight{\@IEEEtrantmpdimenB}{\IEEEarxivheaderstyle HT}\relax
\advance\@IEEEtrantmpdimenA by -1\@IEEEtrantmpdimenB
\setbox\@IEEEboxX=\hbox{\relax
\raisebox{\@IEEEtrantmpdimenA}[0pt][0pt]{\parbox[t]{\textwidth}{\centering
\IEEEarxivheaderstyle\IEEEarxivheader}}}\relax
\wd\@IEEEboxX=0pt\relax
\ht\@IEEEboxX=0pt\relax
\dp\@IEEEboxX=0pt\relax
\box\@IEEEboxX\relax}

\def\IEEEarxivfooterstyle{\normalfont\footnotesize}

\def\IEEEdoarxivfooter#1{\@IEEEtrantmpdimenA\IEEEfoottobottomofpage\relax
\advance\@IEEEtrantmpdimenA by \IEEEarxivheadfootoffset\relax
\settodepth{\@IEEEtrantmpdimenB}{\IEEEarxivheaderstyle gjpqy}\relax
\advance\@IEEEtrantmpdimenA by 1\@IEEEtrantmpdimenB
\setbox\@IEEEboxX=\hbox{\hskip#1\hskip -1in\relax
\raisebox{\@IEEEtrantmpdimenA}[0pt][0pt]{\parbox[b]{\paperwidth}{\centering
\IEEEarxivfooterstyle\IEEEarxivfooter}}}\relax
\wd\@IEEEboxX=0pt\relax
\ht\@IEEEboxX=0pt\relax
\dp\@IEEEboxX=0pt\relax
\box\@IEEEboxX\relax}

\def\@IEEEheaderstyle{\normalfont\scriptsize}
\def\@IEEEfooterstyle{\normalfont\scriptsize}

\makeatother

\renewcommand{\IEEEarxivheadfootoffset}{3pt}

\renewcommand{\IEEEarxivheaderstyle}{\normalfont\footnotesize}
\renewcommand{\IEEEarxivfooterstyle}{\normalfont\footnotesize}

\renewcommand{\IEEEarxivheader}{This is the author's version of an article
that has been published in this journal. Changes were made to this version
by the publisher prior to publication.\\
The final version of record is available at 
\url{http://dx.doi.org/10.1109/TASLP.2015.2389622}}
\renewcommand{\IEEEarxivfooter}{Copyright (c) 2015 IEEE. Personal use is
permitted. For any other purposes, permission must be obtained from the
IEEE by emailing pubs-permissions@ieee.org.}

\usepackage[pdftex]{graphicx}
\DeclareGraphicsExtensions{.pdf,.jpeg,.png}
\usepackage[encapsulated]{CJK}
\usepackage{ucs}
\usepackage[utf8x]{inputenc}
\newcommand{\cntext}[1]{\begin{CJK}{UTF8}{gbsn}#1\end{CJK}}
\usepackage{url}

\newcommand\MYhyperrefoptions{hypertexnames=true, bookmarks=true,
bookmarksnumbered=true, pdfpagemode={UseOutlines}, plainpages=false,
pdfpagelabels=true, colorlinks=true, linkcolor={black},
citecolor={black}, urlcolor={blue}}
\ifCLASSINFOpdf
\usepackage[\MYhyperrefoptions,breaklinks=true,pdftex]{hyperref}
\else
\usepackage[\MYhyperrefoptions,dvips]{hyperref}
\usepackage{breakurl}
\fi

\usepackage{amsmath}
\begin{document}
%
\title{Syntactic and Semantic Features For Code-Switching Factored Language Models}
%
%
%

\author{Heike Adel,
        Ngoc Thang Vu,
        Katrin Kirchhoff,
        Dominic Telaar,
        and Tanja Schultz

\thanks{Copyright (c) 2013 IEEE. Personal use of this material is permitted. However, permission to use this material for any other purposes must be obtained from the IEEE by sending a request to pubs-permissions@ieee.org.}}

%
%

\markboth{IEEE/ACM Transactions on Audio, Speech and Language Processing}%
{IEEE/ACM TRANSACTIONS ON AUDIO, SPEECH AND LANGUAGE PROCESSING}
%



\maketitle
\begin{abstract}
This paper presents our latest investigations on different features for factored language models for Code-Switching speech and their effect on automatic speech recognition (ASR) performance. We focus on syntactic and semantic features which can be extracted from Code-Switching text data and integrate them into factored language models. Different possible factors, such as words, part-of-speech tags, Brown word clusters, open class words and clusters of open class word embeddings are explored. The experimental results reveal that Brown word clusters, part-of-speech tags and open-class words are the most effective at reducing the perplexity of factored language models on the Mandarin-English Code-Switching corpus SEAME. In ASR experiments, the model containing Brown word clusters and part-of-speech tags and the model also including clusters of open class word embeddings yield the best mixed error rate results. In summary, the best language model can significantly reduce the perplexity on the SEAME evaluation set by up to 10.8\% relative and the mixed error rate by up to 3.4\% relative.
\end{abstract}
\section{Introduction}
\label{sec:intro}
The term Code-Switching (CS) denotes speech with more than one language. 
Speakers switch the language while they are talking. 
This phenomenon appears in multilingual communities, such as in India, Hong Kong or Singapore. 
Furthermore, it increasingly occurs in formerly monolingual cultures due to the strong growth of globalization. 
In many contexts and domains, speakers switch between their native language and English within their utterances.
This is a challenge for speech recognition systems, which are typically monolingual. 
While there have been promising approaches to handle Code-Switching in the field of acoustic modeling, language modeling (LM) is still a challenge. 
The main reason is a shortage of training data. 
While about 50h of training data might be sufficient for the estimation of acoustic models, the transcriptions of these data are not enough to build reliable LMs.

The main contribution of this paper is the extensive investigation of syntactic and semantic features for language modeling of CS speech.
Not only traditional features like POS tags and Brown clusters are used but also low dimensional word embeddings.
To easily integrate them into the models, we apply factored language models with generalized backoff. 
The features are analyzed in the context of CS language prediction and automatic speech recognition.

The paper is organized as follows:
Section~\ref{sec:relWork} gives a short overview of related works.
In Section~\ref{sec:corpus}, we describe the data resources which are used in this research work, present different features and analyze them with respect to Code-Switching prediction. 
Section~\ref{sec:FLM:sec:Section1} introduces factored language models.
In Section~\ref{sec:Evaluation:FLM} and \ref{sec:Flm:Dec}, we summarize our most important experiments and results.
The study is concluded in Section~\ref{sec:conclusion}.

\section{Related Work}
\label{sec:relWork}
This section describes previous studies in the field of Code-Switching, language modeling for Code-Switching and factored language models. 
Furthermore, a study of obtaining vector representations for words is presented since they will be used to create additional features in this paper.

In~\cite{syntaxCS3,syntaxCS1,syntaxCS2}, it is observed that Code-Switching occurs at positions in an utterance where it does not violate the syntactic rules of the languages involved.
Code-Switching can be regarded as a speaker dependent phenomenon~\cite{Auer2,slsp2013} but particular CS patterns can also be shared across speakers~\cite{Poplack}. 
It can be observed that part-of-speech (POS) tags may predict CS points more reliably than words themselves. 
The authors of~\cite{CSLM2} predict CS points using several linguistic features, such as word form, language ID, POS tags or the position of the word relative to the phrase. 
The authors of~\cite{CSLM1} compare four different kinds of n-gram lan\-gua\-ge mo\-dels to predict Code-Switching. 
They discover that clustering all foreign words into their POS classes leads to the best performance.
In~\cite{fung01}, the authors propose to integrate the equivalence constraint into language modeling for Mandarin and English CS speech recorded in Hong Kong.
In~\cite{icassp2013}, we extended recurrent neural network language models for CS speech by adding features to the input vector and factorizing the output vector into language classes. 
These models reduce the perplexities and mixed error rates when they are applied to rescore n-best lists.
In contrast to this previous work, we now focus on feature engineering and use a model which can be more efficiently integrated into the first decoding pass than a neural network.

Due to the possibility of integrating various features into factored language models (FLMs), it is possible to handle rich morphology in languages like Arabic~\cite{ga-flm,arabic}. 
In~\cite{acl2013}, we report results of initial experiments with FLMs for CS speech and show that they outperform n-gram language models. 
The best performance is achieved by combining their estimates with recurrent neural network probabilities.
In~\cite{sltu2014}, we present syntactic and semantic features for modeling CS language. This paper is an extension of that study and includes more explanations and analyses, especially for the vector based open class word clusters.

In~\cite{rnnlm-vectors}, the authors explore the linguistic information in the word representation learned by a recurrent neural network. They discover that the network is able to capture both syntactic and semantic regularities. For example, the relationship of the vectors for ``man'' and ``king'' is the same as the relationship of the vectors for ``woman'' and ``queen''. In this paper, these word representations will be used to derive features for FLMs.

\section{Analyses of the Data Corpus with Respect to Possible Factors}
\label{sec:corpus}

This section introduces the corpus used in this work. Furthermore, it presents CS analyses of the text data. Textual features are examined which may trigger language changes.
They are ranked according to their Code-Switching rate (CS rate). The CS rate of each feature $f$ is calculated by its frequency of occurrences preceding CS points divided by its frequency in the entire text:
\begin{equation}
\text{CS rate(f)} = \frac{\text{frequency of f in front of CS points}}{\text{total frequency of f}}
\end{equation}
To provide reliable estimates, only those features are considered whose total frequency exceeds a feature-specific threshold.

\subsection{The SEAME Corpus}
\label{ch:corpus:sec:SEAME}

The corpus used in this thesis is called SEAME (South East Asia Mandarin-English). 
It is a conversational Mandarin-English CS speech corpus recorded by~\cite{SEAME}. 
Originally, it was used for the research project ``Code-Switch'' which was jointly performed by Nanyang Technological University (NTU) and  Karlsruhe Institute of Technology (KIT) from 2009 until 2012.
The corpus consists of 63 hours of audio data and their transcriptions. The audio data were recorded from Singaporean and Malaysian speakers. 
The recordings consist of spontanously spoken interviews and conversations. 
For the task of language modeling and speech recognition, the corpus has been divided into three disjoint sets: training, development (dev) and evaluation (eval) set. 
The data is assigned to the three different sets based on the following criteria: a balanced distribution of gender, speaking style, ratio of Singaporean and Malaysian speakers, ratio of the four language categories, and the duration in each set. 
Table~\ref{csStat} lists the statistics of the SEAME corpus.
\begin{table}[h!]
\caption{\label{csStat}Statistics of the SEAME corpus}
\centering
\begin{tabular}{l|r|r|r}
& Training set & Dev set & Eval set\\
\hline
\# Speakers & 139 & 8 & 8 \\
Duration(hours) & 59.2 & 2.1 & 1.5\\
\# Utterances & 48,040 & 1,943 & 1,029 \\
\# Words & 575,641 & 23,293 & 11,541\\
\end{tabular}
\end{table}

The words can be divided into four language categories: English words ($34.3\%$ of all tokens), Mandarin words ($58.6\%$), particles (Singaporean and Malayan discourse particles, $6.8\%$ of all tokens) and others (other languages, $0.4\%$ of all tokens). 
The language distribution shows that the corpus does not contain a clearly predominant language. 
In total, the corpus contains 9,210 unique English and 7,471 unique Mandarin words. 
The Mandarin character sequences have been segmented into words manually. 
Furthermore, the number of CS points is quite high: On average, there are 2.6 switches per utterance.
Additionally, the duration of the monolingual segments is rather short: More than 82\% of the English segments and 73\% of the Mandarin segments last less than one second.
The average duration of English and Mandarin segments is only 0.67 seconds and 0.81 seconds, respectively. 
This corresponds to an average length of monolingual segments of 1.8 words in English and 3.6 words in Mandarin.

\subsection{Trigger Words}
\label{ch:corpus:sec:trWords}

First, the words occuring in front of CS points are analyzed. 
For this, only those words are considered which appear more than 1,000 times in the text, corresponding to more than 0.2\% of all word tokens. 
By regarding the words with highest CS rates, we notice that in both languages mainly function words (e.g. ``then'', ``but'', ``in'') appear in front of CS points.
Hence in the next step, CS rates of POS tags are examined.

\subsection{Trigger Part-of-Speech Tags}
\label{ch:corpus:sec:trPOS}
Due to the rather small size of the SEAME training text, more general features than words are explored. Since part-of-speech (POS) tags show the syntactical role of the words in the sentence, they can be regarded as syntactic features. 
To be able to investigate POS tags and their distribution in front of CS points, a tagging process needs to be applied first.
\subsubsection{Part-of-speech tagging of Code-Switching speech}
\begin{figure}[h!]
\centering
\includegraphics[width=.35\textwidth]{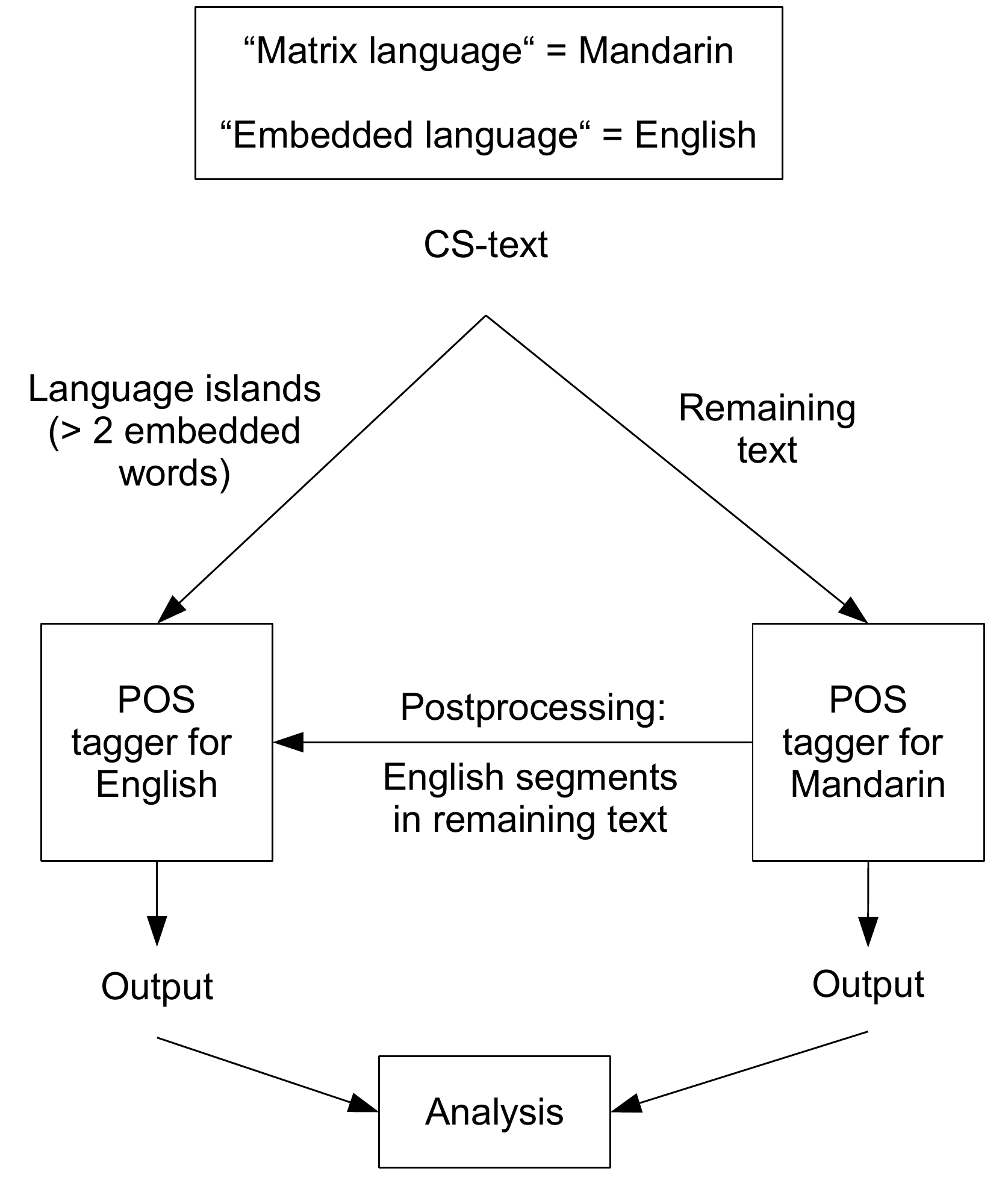}
\caption{Part-of-speech tagger for Code-Switching speech}
\label{pos-tagger}
\end{figure}

For POS tagging of monolingual texts, high quality taggers exist~\cite{Tagger1}. 
However, CS speech contains more than one language. 
Hence, POS tags cannot be determined using a traditional monolingual tagger.
This work uses the POS tagger for CS speech as described in~\cite{CStagger} and illustrated in Figure~\ref{pos-tagger}.

The matrix language is the main language of an utterance, the embedded language is the second language~\cite{matrixLanguage}. 
In the SEAME transcriptions, Mandarin can be determined as the matrix language. 
Three or more consecutive words of the embedded language (English) are called language islands~\cite{matrixLanguage}. All the language islands are passed to the monolingual POS tagger of the embedded language. 
The re\-mai\-ning part is tagged by the monolingual tagger of the matrix language. 
The idea behind this approach is to provide the taggers with as much context as possible. 
Hence, Mandarin segments with only one or two English words are passed to the Mandarin tagger instead of splitting the segments into short monolingual parts. 
This work uses the Stanford log-linear POS tagger for Chinese and English~\cite{Tagger1,Tagger2}. 
The tags are derived from the Penn Treebank POS tag set for Chinese and English~\cite{ChTagset,EnTagset}.\\
However, an analysis shows that most English words which are passed to the Mandarin tagger are incorrectly tagged as nouns (instead of as foreign words). 
Hence, a post-processing step is added to the tagging process to avoid subsequent errors in the determination of trigger POS tags: All English words which do not belong to language islands are selected and passed to the English POS tagger. 
The resulting tags replace the ones obtained from the Mandarin tagger. 
In this step, the English tagger does not get any substantial context (at most one word context) but it is assumed that, nevertheless, its estimates are more appropriate than the estimates of the Mandarin tagger.

\subsubsection{Part-of-speech tag analysis}
In this experiment, we find that
CS points from Mandarin to English are primarily triggered by determiners, while CS from English to Mandarin mainly happens after verbs and nouns. 
This seems reasonable since it is possible that a speaker switches to English for the noun and immediately afterwards back to Mandarin. 

\subsection{Trigger Brown Word Cluster}
\label{ch:corpus:sec:cluster}
The main disadvantage of the POS tags described in the previous section is the lack of evaluation material. 
Since no reference (i.e., correct tagging) for the CS corpus exists, the correctness of the POS tags derived from the tagging process cannot be measured. 
Nevertheless, clustering the words into higher level classes seems to be promising due to the rather small size of the corpus. 
If, for instance, the word ``Monday'' occurs only once and the word ``Tuesday'' occurs once, too, then the class ``days'' occurs at least twice. 
Hence, probabilities might be better estimated for fewer classes than for many different words. 
Therefore, the unsupervised clustering method by Brown et al.~\cite{browncluster} is applied. 
In contrast to POS tags, Brown clusters (Br) are based on word contributions in a text and are, therefore, probably more robust in the case of Code-Switching. 
This clustering method is implemented in the SRILM toolkit~\cite{sri}. 
It uses statistical bigram information to assign words to classes. 
Given the number of classes $C$, the $C$ most frequent words are assigned to their own classes. 
Then, successively, the next most frequent word is selected and assigned to a new class. 
Afterwards, two of the classes are merged. 
The classes to be merged are selected to minimize the average mutual information loss. 
The average mutual information of two classes $c_1$ and $c_2$ is defined as follows~\cite{browncluster}:
\begin{equation}
I(c_1, c_2) = \sum_{c_1 c_2} P(c_1c_2)log\frac{P(c_2 | c_1)}{P(c_2)}
\end{equation}
The sequence $c_1c_2$ denotes that class $c_1$ directly preceeds class $c_2$ in the training text.\\
The resulting classes can be viewed as syntactico-semantic features~\cite{Haffari}.\\
The CS rates observed for Brown word clusters are substantially higher than those of the previous two analyses (ranging up to 73\%, see Table~\ref{trigger-sum}), especially for language changes from English to Mandarin. 
While the rates for a language change from Mandarin to English are higher than 50\% for only two classes, seven classes provide higher rates for a language change in the opposite direction. 
Although the classes have been obtained on the whole training text and, therefore, contain both Mandarin and English words, there is only one class which triggers Code-Switching in both directions. 
It is notable that the classes with the highest CS rates contain more words of the second language than of the one in which they have a trigger function. Hence, it can be speculated that if a foreign word of those syntactical classes is used, it more probably triggers a language change than other words.\\
Due to the higher CS rates compared to the other trigger features, Brown word clusters have a high potential to provide valuable information for language modeling.

\subsection{Open Class Words and Word Vector Based Clusters}
\label{ch:corpus:sec:oc}
While POS tags assign words to classes according to their syntactic function, the algorithm by Brown et al. clusters words based on distributional similarities since it uses bigram counts of the training text. 
In the next step, a different kind of semantic features is investigated. 
Since the SEAME corpus does not contain any semantic tagging, this  paper focuses on open class words and clusters of open class words. 
As described in \cite{rnnlm-vectors}, neural networks are able to learn semantic similarities among words. 
Since those words and similarities are represented as vectors in a continuous space, they can be clustered using vector clustering algorithms, such as k-means or spectral clustering methods. 
Those methods will be described in the following subsections after exploring the usage of open class words as trigger events.

\subsubsection{Trigger open class words}
\label{ch:corpus:sub:oc}
Typically, words can be categorized into closed class words (function words) and open class words (content words). 
Closed class words specify grammatical relations rather than semantic meaning. Examples are conjugations, prepositions and determiners. The class of these words is called ``closed'' since their number is finite and typically no new words are added to them. On the other hand, open class (oc) words express meaning, such as ideas, concepts or attributes. Their class is called ``open'' since it can be extended with new words, such as ``Bollywood''. It contains, for example, nouns, verbs, adjectives and adverbs~\cite{ocDef}.\\
Since open class words carry the meaning of sentences, they can be used to determine the topic of a current utterance. For both languages, English and Mandarin, lists of function words are obtained on the Internet~\cite{EngFunction,ManFunction} and for each word, the preceding open class word is used as a factor.  
Since the CS text contains about 335k open class words, only those open class words with more than 600 occurences in the text are regarded in the CS rate analysis. 
This corresponds to about 0.2\% of all open class words and is, therefore, comparable to the threshold of the trigger words (see Section~\ref{ch:corpus:sec:trWords}).
\\Compared to the trigger words, the CS rates of the open class words do not seem to be promising to predict Code-Switching from Mandarin to English (they are below 35\%). It is notable that for all language changes, Mandarin words were the preceding open class words in most of the cases. There are also open class words which appear in front of CS points in both directions. 

\subsubsection{Trigger open class word clusters}
\begin{figure}[h!]\centering
\includegraphics[width=.25\textwidth]{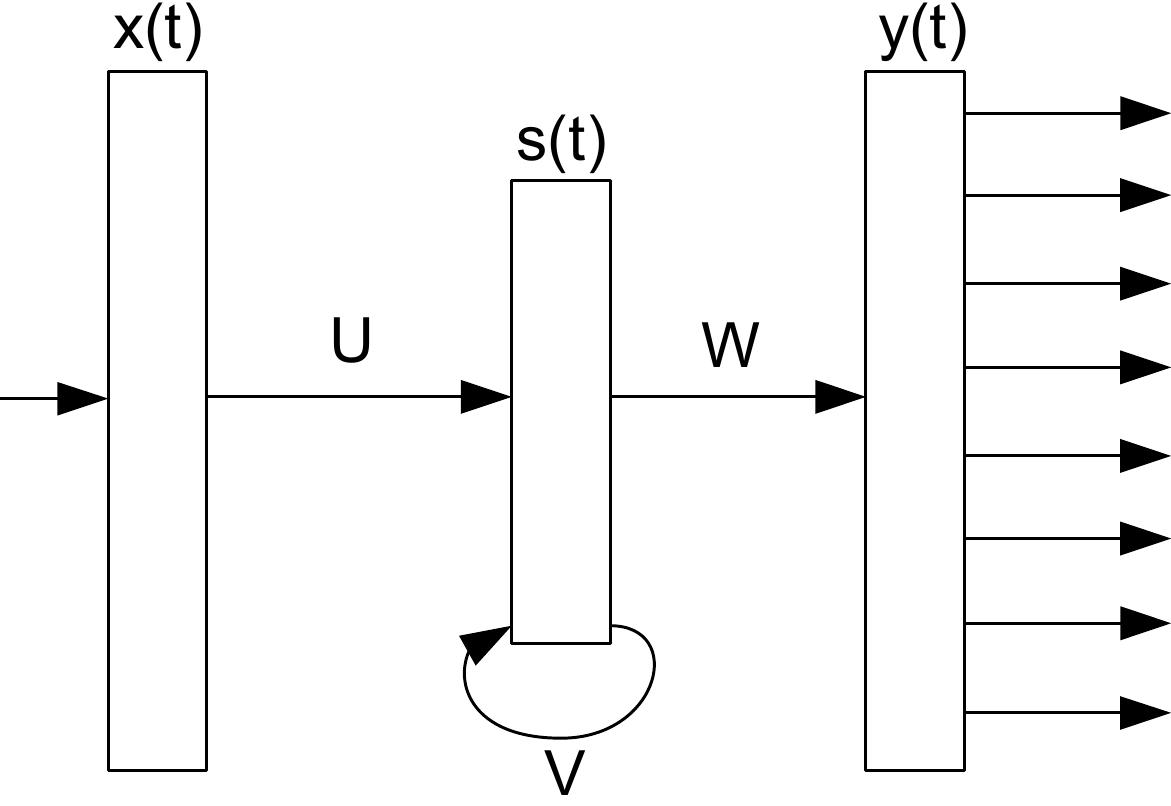}
\caption{Illustration of the components of an RNNLM~\cite{rnnlmTool}}
\label{fig:rnnlm-basics}
\end{figure}

Since the CS rates of open class words are rather low, the implication of clustering them is investigated in the following paragraphs.
In this research, k-means~\cite{kmeans} and spectral clustering~\cite{graclus} are applied to word embeddings extracted from recurrent neural network language models (RNNLMs).
In order to create semantic clusters, only open class words are taken into account because only those are considered to contain meaning (see Section~\ref{ch:corpus:sub:oc}). To increase the number of training examples, two monolingual texts are created: The English text is based on English Gigaword data (fifth edition, corpus number: LDC2011T07) and several more corpora (ACL/DCI (LDC93T1), American National Corpus (LDC2005T35), ACQUAINT corpus (LDC2002T31)).
The Chinese text is based on Chinese Gigaword data (fifth edition, corpus number: LDC2011T13). All the texts mainly contain news articles. In each text, function words are deleted and only lines with a high coverage of the SEAME vocabulary are selected.
In particular, the resulting texts consist of about 630k Chinese words and 654k English words. They are divided into a training and a development set (at a ratio of 10 to 1). Based on these texts, two monolingual RNNLMs are trained using the toolkit provided in~\cite{rnnlm1}.\\
An RNNLM consists of three layers (see Figure~\ref{fig:rnnlm-basics}): an input layer, a hidden layer and an output layer. The input of the hidden layer does not only depend on the input layer but also on the hidden layer of the previous time step. This is why the network is called ``recurrent''. The input layer is formed by a vector of the size of the vocabulary. 
A word in the training text is represented by a vector containing ``$1$'' at the word index position and ``$0$'' in all the other entries. Similar to the input vector, the output vector consists of one entry for each word of the vocabulary. 
It provides a probability distribution for the next word in the text. 
For training, backpropagation through time~\cite{bptt,rnnlm2} is applied. 
\\After training, embeddings for the words can be found in the weight matrix connecting the input and the hidden layer~\cite{rnnlm-vectors}. 
For the creation of syntactic and semantic clusters, we extract all the vectors whose corresponding words are part of the SEAME vocabulary and cluster them using k-means and spectral clustering. 

\subsubsection{Open class word clusters analysis}
\label{ch:corpus::sub:ocanalysis}
Similar to the CS rates of the open class words (see Section~\ref{ch:corpus:sub:oc}), most clusters preceding a CS point from English to Mandarin are Mandarin clusters. For the opposite direction, however, two English clusters are now among the clusters preceding CS points most often. 
The CS rates of the open class word clusters improve the CS rates of the open class words slightly.
In total, open class words (clusters) better predict switches from EN to MAN than POS tags.

\subsection{Summary: Comparison of the Trigger Features}
To sum up, for Code-Switching from English to Mandarin, Brown word clusters provide the highest CS rates. For Code-Switching from Mandarin to English, the best CS rates are obtained with trigger words. This motivates the combination of different features into one FLM.
Table~\ref{trigger-sum} provides an overview of the CS rates of the different trigger features. 
\begin{table}[h!t]
\centering
\caption{Overview of the CS rates of different trigger features}
\begin{tabular}{l|c|c}
Feature & CS rate: MAN to EN & CS rate: EN to MAN \\
\hline
Words & $\le$ \textbf{53.43\%} & $\le$ 56.25\% \\
Part-of-speech tags & $\le$ 43.13\% & $\le$ 47.78\% \\
Brown word clusters & $\le$ 52.73\% & $\le$ \textbf{72.67\%} \\
Open class words & $\le$ 33.33\% & $\le$ 54.53\%\\
Open class word clusters & $\le$ 34.44\% & $\le$ 56.66\%\\
\end{tabular}
\label{trigger-sum}
\end{table}
\\In average, the Brown word clusters seem to be the most promising features for the prediction of CS points.

\section{Use of Factored Language Models}
\label{sec:FLM:sec:Section1}

\subsection{Language modeling}
Language models calculate the probability of a word sequence $W$~\cite{lm}:
\begin{equation}
\begin{split}
P(W) = P(w_1) \cdot P(w_2 | w_1) \cdot ... \cdot P(w_n | w_1 w_2 ... w_{n-1})\\
 = \prod_{i = 1}^{n}P(w_i | w_1, w_2, ..., w_{i-1})
\end{split}
\end{equation} 
N-gram language models limit the context for this computation to the previous $n-1$ words and the current (next) word.
\begin{equation}
P(s) = \prod_{i = 1}^{k}P(w_i | w_{i-n+1}^{i-1})
\end{equation}
The probabilities are estimated based on counts of words and contexts in a training text. 
Due to their computational efficiency, mainly n-gram models are used in the field of speech recognition.

\subsection{Factored language models}
Factored language models (FLMs) consider vectors of features (e.g. words, morphological classes, word stems or clusters)~\cite{flm-tut,flm}. The following equation expresses that a word $w_t$ is regarded as a collection of factors $f_t^1, f_t^2, ... f_t^K$: 
\begin{equation}
w_t \equiv \left\{f_t^1, f_t^2, ... f_t^K\right\} = f^{1:K}_t
\end{equation}
Similar to n-gram language models, the probability for the next word is computed based on counts of factor context occurrences 
in the training text. However instead of a word context $w_{i-n+1} ... w_{i-1}$, a pre-defined factor context is used. This context is one of the main design choices when building a factored language model. It could look as follows: $f_{t-2}^1 f_{t-1}^1 f_{t-1}^2$. This example context would lead to the following equation for calculating the probability of word $w$:
\begin{equation}
P(w | context) = \frac{Count(f_{t-2}^1 f_{t-1}^1 f_{t-1}^2 w)}{Count(f_{t-2}^1 f_{t-1}^1 f_{t-1}^2)}
\end{equation}
The advantage of regarding factor contexts instead of word contexts is that usually there are fewer different factors than different words. 
Hence, the coverage of factor contexts in training texts may be greater than that of equally long word contexts. 
This is especially important for short training texts. 
Nevertheless, it is unlikely to see all factor context combinations. 
In the case of unseen contexts, generalized backoff is performed. 
This means that some of the factors in the context are dropped. 
For each omitted factor, a backoff result is calculated. 
If there is more than one factor which can be dropped, the backoff results are combined, for instance using their average, their sum, or their product.

\section{FLMs: Perplexity Results}
\label{sec:Evaluation:FLM}

For the task of Code-Switching, a variety of FLMs is trained and evaluated with different conditioning factors. 
The backoff paths and smoothing options are chosen for each FLM individually in order to minimize its perplexity on the development set. For each feature combination, an initial set of parameters is obtained using a genetic algorithm~\cite{ga-flm}. Its results are then improved manually by changing single parameters. 
According to the analyses in Section~\ref{sec:corpus}, the following factors are investigated: words, POS tags, Brown word clusters, open class words and open class word clusters.\footnote{Note that the Brown word clusters are bilingual clusters since they are created on the CS training text while the POS tags group the words into monolingual classes. For the open class word clusters, bilingual classes led to better results than monolingual classes. This is further described in Section~\ref{ch:Evaluation:sec:Experiments:sub:ocCluster}.}\\ 
For each feature (besides words), an FLM which uses only words and this feature as conditioning factors is built. Furthermore, FLMs are created which combine different features.\\
The following subsections describe the perplexity results for the different FLMs. 

\subsection{Factors: part-of-speech tags and language information}
\begin{table}[h!t]
\centering
\caption{PPL of FLMs with POS tags and LID}
\begin{tabular}{l|c|c}
Model & PPL dev & PPL eval\\
\hline
Baseline (3-gram) & 268.39 & 282.86\\
\hline
POS & 260.70 & 267.86\\
LID & 263.24 & 267.63\\
POS + LID & \textbf{257.62} &  \textbf{264.20}\\
\end{tabular}
\label{FLM1}
\end{table}
First, the factor POS tag is explored. A language model containing the last word and the two last POS tags as conditioning factors is trained. This choice of conditioning factors is obtained by the genetic algorithm and maintained during the manual optimization. 
Second, a language model using only language identifiers (LID) and words as features is trained. 
Finally, the factors words, POS tags and language information are combined. The perplexity results show that they provide complementary information which helps to improve the language model predictions. Table~\ref{FLM1} summarizes the results of these experiments.

\subsection{Factor: Brown word clusters (BrC)}
\label{ch:Evaluation:sec:Experiments:sub:wordcluster}
In the next experiments, Brown word clusters are explored. As described in Section~\ref{ch:corpus:sec:cluster}, the SRILM toolkit~\cite{sri} is used to obtain the clusters. 
To determine the number of Brown word clusters, FLMs are trained using words and Brown word clusters of different sizes as features. 
Since the clusters should help to improve the CS language models, their perplexity on the SEAME development set is calculated.
Class numbers in the range of 50 to 100 lead to the best performance. A class size of 70 is chosen for the following experiments.

Brown word clusters with 70 classes are first investigated as the only factor besides words. Then, they are combined with the factors POS tags and LID. Table~\ref{FLM2b} presents the experimental results.
The results show that the combination of Brown word clusters and POS tags leads to the best word prediction results on the CS text. The additional integration of LID does not improve the results.
\begin{table}[h!t]
\centering
\caption{PPL of FLMs with Brown word clusters, POS tags and LID}
\begin{tabular}{l|c|c}
Model & PPL dev & PPL eval\\
\hline
Baseline (3-gram) & 268.39 & 282.86\\
\hline
BrC & 257.17 & 265.50\\
BrC + POS & \textbf{249.00} & \textbf{255.34}\\
BrC + LID & 260.39 & 268.71\\
BrC + POS + LID & 251.39 & 259.05\\
\end{tabular}
\label{FLM2b}
\end{table}

\subsection{Factor: open class words}
\label{ch:Evaluation:sec:Experiments:sub:oc}
First, three different ways of assigning an open class word factor to words are compared: For each word, the previous open class word is determined and added as a factor. At the beginning of each sentence, this factor can be reset to an unknown tag (1). This is based on the idea that in each sentence, there might be different topics and open class words and, therefore, a reset might be necessary. This approach will be referred to as ``last open class word per sentence''. Another possibility is to keep the previous open class word over sentence boundaries but to reset it at every speaker change (2). This is reasonable since the same speaker may talk about only a limited number of topics and, therefore, use similar open class words. Another speaker, however, may address a different subject. Although the corpus contains conversations, there is no information about which speakers talk with each other about the same topic. Hence, the open class words used by different speakers may be different, too. This approach is called ``last open class word per speaker'' in the following table. The last approach tries to generalize the open class words into topics. For each speaker, the most frequent open class word in a window of the previous $n$ open class words is used as a factor (3). If several open class words have the same frequency, the most recent one is chosen. This method will be referred to as ``most frequent open class word in window''. Tables~\ref{unclustered1} and~\ref{unclustered2} provide the perplexity results when FLMs based on these different approaches are built. Table~\ref{unclustered1} shows the results if only words and open class words are used as factors while Table~\ref{unclustered2} provides an overview of the results if POS tags and Brown word clusters are also integrated into the models. Based on the results of the previous experiments, language information tags are not used as additional factors.
\begin{table}[h!t]
\centering
\caption{PPL of FLMs with words and open class words}
\begin{tabular}{l|c|c}
Approach & PPL dev & PPL eval \\
\hline
Baseline (3-gram) & \textbf{268.39} & 282.86\\
\hline
(1) Last open class word per sentence & 278.33 & \textbf{279.60}\\
\hline
(2) Last open class word per speaker & 278.12 & 281.31\\
\hline
(3) Most frequent open class word in window: & & \\
\hspace*{1cm}Window size: unlimited & 296.52 & 299.35\\
\hspace*{1cm}Window size: 10 & 287.19 & 290.80 \\
\hspace*{1cm}Window size: 5 & 284.19 & 288.95\\
\end{tabular}
\label{unclustered1}
\end{table}
\begin{table}[h!t]
\centering
\caption{PPL of FLMs with words, Brown word clusters, POS tags and open class words}
\begin{tabular}{l|c|c}
Approach & PPL dev & PPL eval \\
\hline
Baseline (3-gram) & 268.39 & 282.86\\
\hline
(1) Last open class word per sentence &  247.64 & \textbf{251.73}\\
\hline
(2) Last open class word per speaker &  \textbf{247.18}& 252.37\\
\hline
(3) Most frequent open class word in window:  & &\\
\hspace*{1cm}Window size: unlimited &   262.13 & 263.12\\
\hspace*{1cm}Window size: 10 & 254.31 & 260.68\\
\hspace*{1cm}Window size: 5 &  252.40 & 259.25\\
\end{tabular}
\label{unclustered2}
\end{table}
\\The results show that changing the realization of the open class word feature more often (in the case of smaller window sizes) leads to better results. The model ``last open class word per speaker'' corresponds to the model ``most frequent open class word in window'' with a window size of 1. It results in the lowest perplexities. Resetting the open class word after each sentence leads to only slightly worse results. It is notable that the perplexities are reduced very much by combining Brown word clusters, POS tags and open class words. The reason for this could be the backoff. The FLM with words and open class words needs to backoff to (open class) words while the other FLM can also backoff to word clusters. Since there are fewer clusters than words in the text, specific cluster combinations may appear more often than specific word combinations. 
For the following experiments, the last open class word per speaker is used as the realization of the open class word factor. In order to improve the so-far best language model, the FLMs of the following experiments use the factors words, POS tags and Brown word clusters in addition to open class words (clusters).

\subsection{Factor: open class word clusters}
\label{ch:Evaluation:sec:Experiments:sub:ocCluster}
In the following experiments, the open class words are grouped using different clustering methods. For this, both the CS training text and the monolingual texts as described in Section~\ref{ch:corpus::sub:ocanalysis} are used. If the clusters are based on the monolingual texts, not every word from the CS text is covered by the classes. In this case, the factor is formed by the word itself instead of a class. The results are evaluated regarding the perplexities of the FLMs. For all the language models, the same parameters are used in order to ensure comparability. Table~\ref{clustered} summarizes all the results. The different approaches are explained in the following paragraphs.
\begin{table}[h!t]
\caption{PPL of FLMs with words, part-of-speech tags, Brown word clusters and different open class word clusters}
\centering
\begin{tabular}{l|c|c}
Approach & PPL dev & PPL eval\\
\hline
Unclustered & \textbf{247.18} & \textbf{252.37}\\
\hline
(1) Brown clusters  & 254.23& 260.29\\
\hline
(2) Oc BrC 2k CS classes & 248.07 & 253.01\\
(2) Oc BrC 4k CS classes & 247.52 & 252.44\\
(2) Oc BrC 6k CS classes &  247.47 & 252.53\\
(2) Oc BrC 1k EN + 1k MAN classes  & 248.40 & 253.78\\
(2) Oc BrC 2k EN + 2k MAN classes  & 247.89 & 252.84\\
(2) Oc BrC 3k EN + 3k MAN classes &  247.56 & 252.61\\
\hline
(3) K-means clusters 1k EN + 1k MAN classes  & 249.13& 254.13\\
(3) K-means clusters 2k EN + 2k MAN classes &  248.26&253.18\\
(3) K-means clusters 3k EN + 3k MAN classes  & 247.97&252.81\\
\hline
(4) Spectral clusters 1k EN + 1k MAN classes  & 248.93&254.07\\
(4) Spectral clusters 2k EN + 2k MAN classes  & 248.31&252.94\\
(4) Spectral clusters 3k EN + 3k MAN classes & 248.02& 252.69\\
\hline
(5) Spectral clusters 1k CS classes & 251.61&255.87\\
(5) Spectral clusters 2k CS classes  & 250.53&254.65\\
(5) Spectral clusters 3k CS classes  & 249.97&254.65\\
\hline
(6) ML spectral clusters 250 classes  & 249.04& 254.61\\
(6) ML spectral clusters 500 classes  & 248.12& 253.35\\
(6) ML spectral clusters 800 classes  & 247.24& 252.60\\
\end{tabular}
\label{clustered}
\end{table}
\\As a first approach, the Brown word clusters as described in Section~\ref{ch:Evaluation:sec:Experiments:sub:wordcluster} are used instead of the open class words themselves (1). Hence, the number of possible realizations of the open class factor is limited to 70. The factor values are more general compared to using the open class words themselves. This approach is called ``Brown clusters'' in the table. The result shows that although the Brown word cluster of the previous word has added useful information to the language modeling process, the Brown cluster of the preceding open class word seems to rather add more confusability since the perplexity is increased. An explanation could be that the Brown clusters have been trained on the whole training text including function words.\\
Second, the Brown clustering algorithm is applied only to the open class words (2). For this, all function words are deleted from the CS text. Again, different cluster sizes are explored. Furthermore, the monolingual open class texts as described in Section~\ref{ch:corpus::sub:ocanalysis} are used to cluster the English and Mandarin words individually. Thus, the models called ``oc Brown clusters CS classes'' consist of classes containing both English and Mandarin words while the models ``oc Brown clusters EN + MAN classes'' include separate classes for English and Mandarin words. The number of the CS classes is chosen to be the same as the sum of EN classes and MAN classes. The results show that the models perform better than ``(1) Brown clusters''. Furthermore, it can be seen that the performance is improved with an increasing number of classes.\\
The clusters labeled with (1) and (2) are distribution based clusters. The following experiments explore semantic clusters. As described in Section~\ref{ch:corpus::sub:ocanalysis}, two RNNLMs are trained on English and Mandarin monolingual texts, respectively. Then, the word vectors which are stored in the weight matrizes between the input and the hidden layer are extracted and clustered. Those clusters are, then, used as features in the FLMs. All the open class words which are not covered by the classes because they do not appear in the monolingual texts, stay the same. This affects 5,271 different words (32.21\% of all open class words) which occur in total 73,478 times (12.76\% of all tokens) in the training text.
Since the two different monolingual networks learn different word representations, Mandarin words which are similar to English words might not be assigned to similar vectors and as a result, not to the same class. Hence, monolingual clusters are computed for English and Mandarin.\\
First, k-means is used for clustering (3). The results are listed with the name ``k-means clusters''. Since the previous results showed that an increase of the number of classes leads to better performance, rather high class numbers are used in the experiments. Indeed, the results show again that the perplexity decreases if the number of classes is raised. A possible evaluation of the clustering quality is the calculation of inter-cluster and intra-cluster variances. Inter-cluster variance denotes the distance of different clusters while intra-cluster variance shows how compact a cluster is. Based on~\cite{evalKmeans}, the variances are calculated as shown in the following equations.
\begin{equation}
\begin{split}
intra = \frac{1}{N}\sum_{i=1}^{k}\sum_{x \in c_i} |x - \mu^{(c_i)}|^2\\
inter = min(|\mu^{(c_i)} - \mu^{(c_j)}|^2), i = 1..k-1, j = i+1..k\\
\end{split}
\end{equation}
The term $\mu^{(c_i)}$ denotes the mean vector of class $c_i$, $k$ the number of classes and $N$ the number of vectors.
Furthermore, a validity ratio is computed as follows:
\begin{equation}
ratio = \frac{intra}{inter}
\end{equation}
Since the intra-cluster variance should be minimized while the inter-cluster variance should be maximized, lower ratios correspond to better clustering results. Table~\ref{fisherratio} provides the variances and ratios for the k-means clusters of different sizes.
\begin{table}[h!]
\caption{Intra-class variances, inter-class variances and validity ratios for different k-means cluster sizes}
\centering
\begin{tabular}{l|c|c|c}
Clustering & intra-class variance & inter-class variance &ratio\\
\hline
1000 EN classes & 0.0911 &0.0088&10.39\\
2000 EN classes & 0.0505&\textbf{0.0105}&4.82\\
3000 EN classes & \textbf{0.0277}&0.0104&\textbf{2.67}\\
\hline
1000 MAN classes & 0.0804&\textbf{0.0020}&40.54\\
2000 MAN classes & 0.0454&0.0011&40.38\\
3000 MAN classes & \textbf{0.0262}&0.0016&\textbf{15.99}\\
\end{tabular}
\label{fisherratio}
\end{table}
\\The ratios of the different k-means clusters show that the clustering quality is increased with a larger amount of classes. While the intra-class variances are improved in all cases, the inter-class variances are not always raised. This shows that a higher class number leads to more compact classes which are not necessarily better separated from each other.\\
Since the word classes might not be linearly separable, spectral clustering is applied to the word vectors in the next step (4). The results are called ``spectral clusters''. To provide comparability, the same class sizes are used as for k-means. The results show that spectral clustering leads to better perplexities than k-means clustering although the difference is very small.\\
Table~\ref{cluster-ex} provides examples for words which are grouped into one class using spectral clustering with 2000 classes.
\begin{table}[h!]
\caption{Each column represents an example of the classes obtained by spectral clustering with 2000 classes}
\begin{tabular}{c|c|c|c|c}
friday & august & championships & brazilian & gym \\
thursday & books & elephants & german &  swim\\
tuesday & june & olympics & italian & ski  \\
wednesday & december & stadium & swiss & skiing \\
\end{tabular}
\label{cluster-ex}
\end{table}
\\In order to see how much additional information monolingual texts provide compared to the CS training text, a third RNNLM is trained using the open class words of the CS text as input (5). Its word vectors are clustered using spectral clustering again. The table entries ``spectral clusters CS classes'' provide the results of this experiment. These models perform worse than the models with clusters based on the monolingual texts. The reason for this may lie in the clustering results. The CS spectral clusters do not group semantically similar words into the same class. The word ``august'', for example, is grouped with the English words ``lag'' and ``subjects'' and the Mandarin words ``\cntext{墨}'' (ink), ``\cntext{层}'' (layer), ``\cntext{成了}'' (became), ``\cntext{断绝}'' (sever) and ``\cntext{用法}'' (usage). Reasons for this may be the small amount of CS training data or the bilinguality of the text.\\
In order to experiment with multilingual spectral clusters, a training text for the RNNLM is created using lines of both the English and the Mandarin texts (6). During training, the hidden layer of the network is reset after each line. Hence, the English and Mandarin words are trained with the same network but separately from each other. This seems to be reasonable since the sentences are extracted from different news texts. Therefore, an English sentence does not depend on the previous Mandarin sentence and vice versa. Again, the resulting word vectors are clustered using spectral clustering. Due to the combination of both languages, the classes will consist of both English and Mandarin words. The results of the FLMs using those classes as features are called ``multilingual (ML) spectral clusters'' in the table. Class sizes beyond 800 classes are not investigated since the 800 classes only contain about 1.17 words on average (mi\-ni\-mum: 1 word, maximum: 4 words). Furthermore, 15,450 words (94.40\% of all distinct open class words) which occur in total 201,210 times (34.95\% of all tokens) in the training text are not covered by the multilingual clusters. For these words, the open class words themselves are used instead of classes as described above.
\\All the clustering experiments could not lead to FLMs superior to the model with unclustered open class words. However, the difference among the perplexity results is not large enough to be able to decide which model performs the best.
The best cluster size seems to be at or even beyond 3000 classes. However, those classes do not contain many words. The classes of the English ``oc Brown clusters'' of size 3000, for example, contain 1 to 9 words per class and on average 5.45 words. The classes of the Mandarin ``oc Brown clusters'' of size 3000 also contain 1 to 9 words per class but on average only 1.89 words. Hence, the difference to unclustered words is rather small. An explanation why open class words seem to outperform open class word classes could be the higher branching factor after a class with many members compared to the branching factor after a single word. This might suppress the positive backoff effect of clusters in this case. 
\\Since the sixth approach (multilingual spectral clustering) with 800 classes performed best in terms of perplexity on the dev set, this model will be used in the decoding experiments. It will be referred to as ``open class word clusters''.

\subsubsection{Analysis of open class word clusters}
\begin{figure}[h!t]
\centering
\includegraphics[width=.5\textwidth]{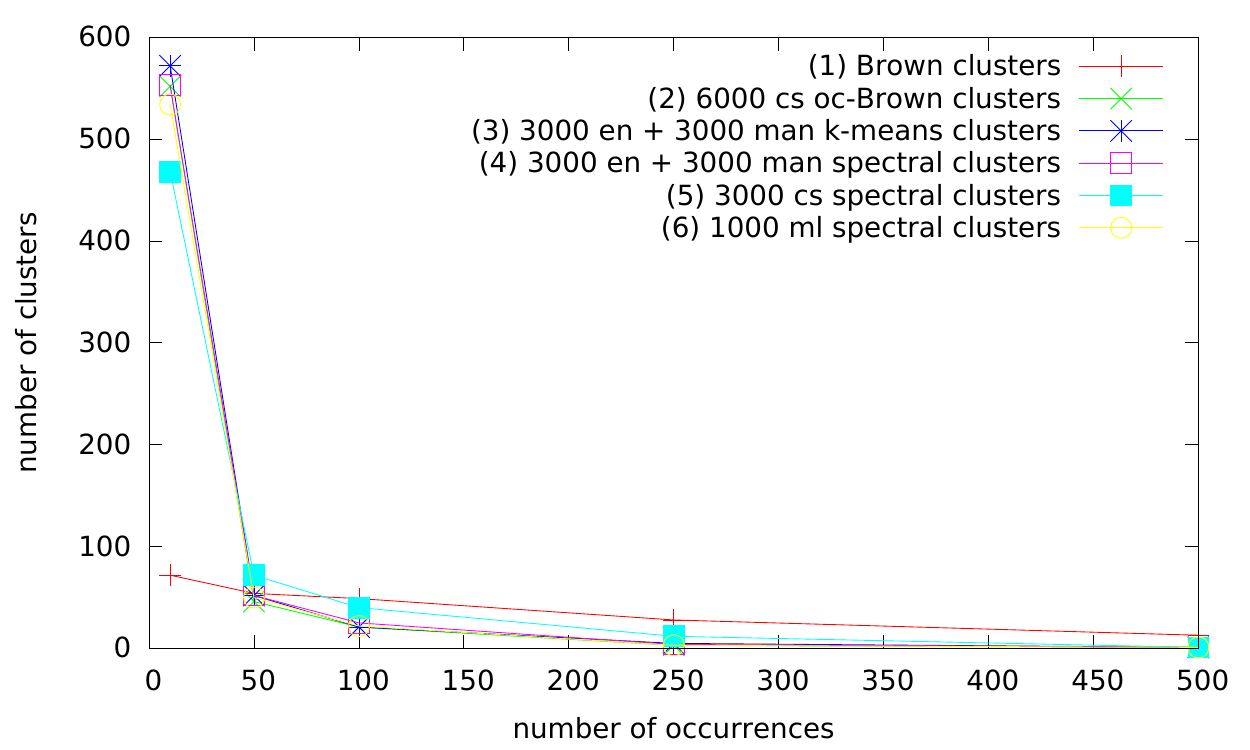}
\caption{Number of clusters occuring certain times in the SEAME development text}
\label{histogram}
\end{figure}

To further evaluate the open class word clusters, an analysis of their distribution is performed. The number of occurrences of each class of the ML spectral clusters with 800 classes in the SEAME development set is counted. Then, it is extracted how many clusters occur more than 10, 50, 100, 250 and 500 times. Figure~\ref{histogram} shows the results. It can be noted that only few clusters occur more than 100 times in the text. 

\subsection{Perplexity results: summary}
\label{ch:Evaluation:sub:FLM-summary}
\begin{figure*}[h!t]
\centering
\includegraphics[width=.7\textwidth]{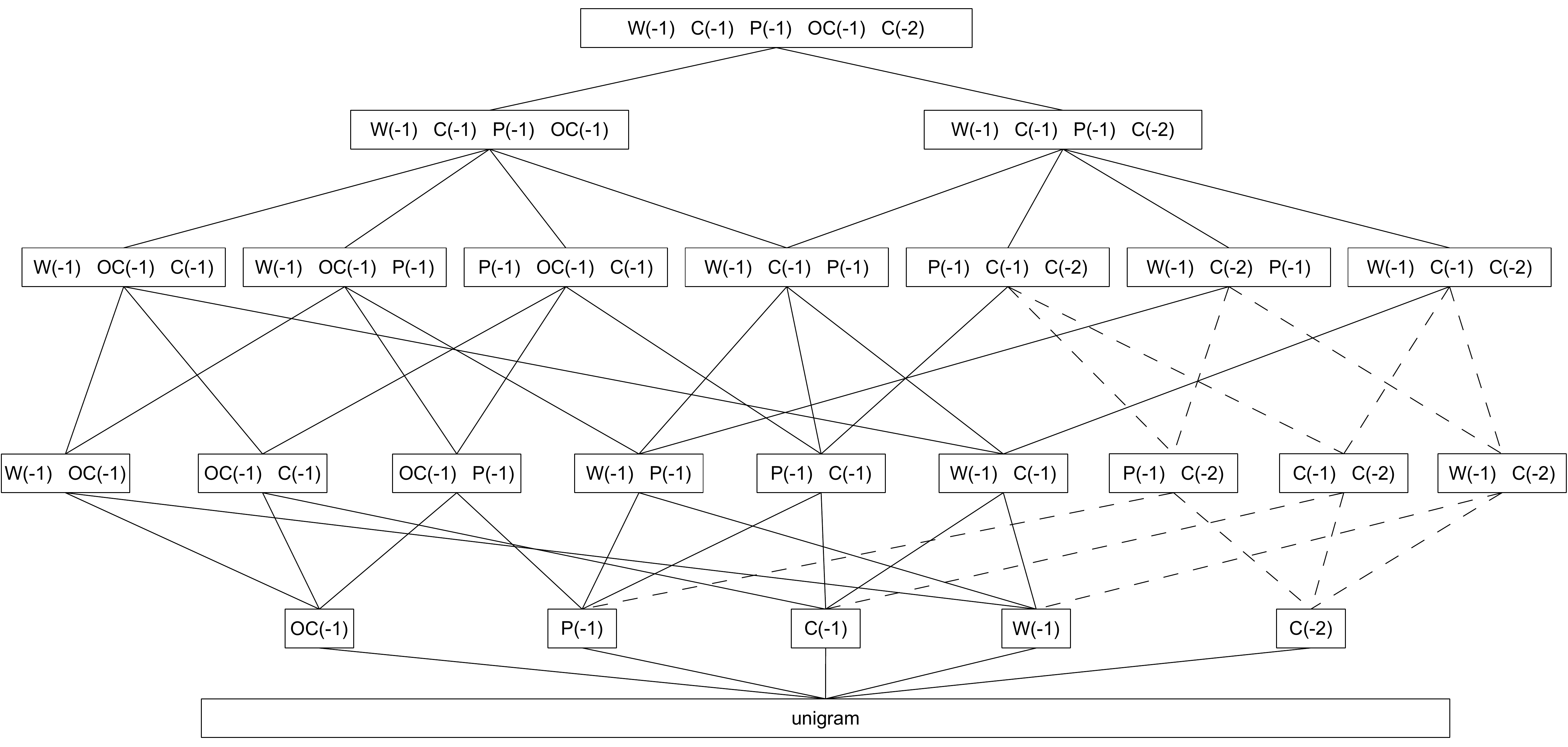}
\caption{The backoff graph of the best FLM (Dashed lines indicate the application of general backoff instead of averaging the results of fixed backoff paths)}
\label{best-bograph}
\end{figure*}

The largest perplexity improvements are obtained by using Brown word clusters (alone and in combination with other features). This corresponds to the observations in Section~\ref{sec:corpus} since Brown word clusters provide the highest CS rates on average. Interestingly, the CS rates of open class word clusters are superior to the rates of open class words but this does not transfer to the perplexity results. A possible explanation is a higher branching factor after clusters in contrast to words.
\\The FLM which performs best in terms of perplexity consists of the factors words, POS tags, Brown word clusters and open class words. Its conditioning factors and backoff paths are shown in Figure~\ref{best-bograph}.
The main idea behind the backoff graph is to first drop either the oldest feature (\texttt{C(-2)}) or the open class word (\texttt{OC(-1)}) in order to continue with a model similar to the second best model (FLM Brown word clusters + POS). 
Afterwards, the results of all possible backoff paths are combined using their average. 
In the case of backoff to one or two factors including the penultimate Brown word cluster (\texttt{C(-2)}), general backoff is applied (indicated by dashed lines in the figure).
A possible explanation is that the penultimate Brown word cluster may be necessary for the prediction of the next word in some but not in all cases. The backoff graph has been investigated experimentally and chosen because of superior results in terms of PPL on the dev set compared to other backoff strategies.\\
Table~\ref{FLM-sum} summarizes the most important results of the experiments conducted with different factors for FLMs. In addition, it provides results and weights for interpolating the FLMs with the baseline 3-gram language model.
\begin{table}[h!t]
\centering
\caption{Summary: PPL of different FLMs, compared to and interpolated with the baseline 3-gram model}
\begin{tabular}{l|c|c}
Model & PPL dev & PPL eval \\
\hline
Baseline (3-gram) & 268.39 & 282.86\\
\hline
FLM POS + LID & 257.62 &  264.20\\
+ CS 3-gram ($w_{FLM}$ = 0.55) & 246.36 & 253.27\\
\hline
FLM BrC + POS & 249.00 & 255.34\\
+ CS 3-gram ($w_{FLM}$ = 0.63) & 241.89 & 248.53\\
\hline
FLM open class words + BrC + POS & 247.18 & 252.37\\
+ CS 3-gram ($w_{FLM}$ = 0.63) & 238.87 & \textbf{245.27}\\
\hline
FLM open class clusters + BrC + POS &  247.24 & 252.60\\
+ CS 3-gram ($w_{FLM}$ = 0.63) & \textbf{238.86} & 245.40
\end{tabular}
\label{FLM-sum}
\end{table}
\\It can be found that interpolating the FLMs with the baseline 3-gram model leads to superior perplexity results in all cases. Except for one model (the FLM using only open class words), the interpolation weight for the FLM is always above 0.5. This shows the high impact of syntactic and semantic features for language modeling for Code-Switching.\\
All factored language models lead to perplexity results which are statistically significantly better than the baseline 3-gram perplexities. The models with Brown word clusters are also significantly superior to the models without. However, the difference between the model with open class words and the best model with open class word clusters cannot be considered statistically significant.

\section{FLMs in the Decoding Process}
\label{sec:Flm:Dec}
\subsection{Using FLM during decoding}
For the ASR experiments, the speaker independent acoustic model and the pronunciation dictionary of the ASR system described in~\cite{CSASR1} are used.
The decoding is performed using the BioKit decoder~\cite{BioKIT}.
During decoding, the baseline 3-gram language model is used for lookahead. 
At every word end, the language model score is obtained by interpolating the FLM and the 3-gram language model. 
The interpolation weight is chosen based on mixed error rate results on the development set. 
For these experiments, the FLM containing only words and POS tags is used. 
For each speaker, the first 50 sentences are decoded. 
This corresponds to more than 20\% of all sentences of the development set. 
This number has been chosen to, on the one hand, achieve reliable results but, on the other hand, reduce computational efforts. 
The experiments reveal that an interpolation weight of 0.45 lead to the lowest mixed error rates. 
Table~\ref{decoding-flm} presents the mixed error rates when the different FLMs are used in the decoding process. 
To be able to compare the mixed error rate results with the perplexities, the perplexity results of the FLMs are presented when they are interpolated with the decoder baseline language model using a weight of 0.45.
The decoding results show that the mixed error rate is not always correlated with the perplexity results. 
However, all the FLMs outperform the traditional 3-gram language model.
\begin{table}[h]
\caption{Mixed error rate and perplexity results for the different FLMs when they are interpolated with the CS 3-gram using an FLM interpolation weight of 0.45}
\centering
\begin{tabular}{l|c|c|c}
Model & MER dev & MER eval  & PPL dev\\
\hline
Decoder baseline 3-gram & 39.96\% & 34.31\% & 292.58\\
\hline
POS & 39.47\% & 33.46\% &250.64\\
POS + LID & 39.66\% & 33.30\% &248.38\\
BrC & 39.45\% & 33.93\% & 249.05\\
BrC + POS & \textbf{39.30\%} & 33.60\% & \textbf{244.62}\\
BrC + POS + LID & 39.39\%  & 33.16\% & 248.64\\
Oc words + BrC + POS& 39.33\%& \textbf{33.15\%} &245.79\\
Oc clusters (ML spectral 800 cl) & \textbf{39.30\%} & 33.16\% & 245.79 \\
+ BrC + POS & & & \\
\end{tabular}
\label{decoding-flm}
\end{table}

\subsection{Analysis of results}
To obtain a better understanding of the advantages of the FLMs, an error analysis is provided in Table~\ref{error_analysis_flm}. The results of the FLMs which lead to the best mixed error rate results on the development set (FLM Brown word clusters + POS tags and FLM Brown word clusters + POS tags + open class word clusters) are compared to the results of the baseline model in detail. Since some of the numbers denote accuracy values and others are error rates, a language model is not always superior if its number is higher (lower). 
\begin{table}
\centering
\caption{Result analysis after decoding with the decoder baseline model and FLM 1 (Brown word clusters + POS tags) or FLM 2 (Brown word clusters + POS tags + oc word clusters)}
\begin{tabular}{l|c|c|c}
& Baseline & FLM 1 & FLM 2 \\
\hline
MER in English segments & 59.40\% & \textbf{57.52\%} & \textbf{56.02\%} \\
MER in Mandarin segments & 36.48\% & \textbf{36.12\%} & \textbf{36.24\%} \\
\hline
Correct words & 64.37\% & 64.39\% & 64.35\% \\
Deletion of English words & 1.65\% & 1.83\% & 1.86\% \\
Deletion of Mandarin words & 5.79\% & 6.32\% & 6.31\% \\
Insertion of English words & 1.09\% & 0.87\% & 0.84\% \\
Insertion of Mandarin words & 2.99\% & 2.62\% & 2.67\% \\
Substitution of EN with EN & 4.87\% & 4.76\% & 4.68\% \\
Substitution of EN with MAN & 4.38\% & 4.41\%  & 4.44\% \\
Substitution of MAN with MAN & 15.30\% & 14.93\% & 14.93\% \\
Substitution of MAN with EN & 2.85\% & 2.50\% & 2.57\% \\
\hline
Word correct after CS & 37.52\% & \textbf{37.80\%} & \textbf{37.61\%} \\
Language correct after CS & 68.23\% & 66.70\% & 66.79\%\\
\end{tabular}
\label{error_analysis_flm}
\end{table}
The FLMs lead to improvements regarding insertion errors and monolingual segments. Furthermore, words at CS points are recognized more robustly.\\
The mixed error rate results obtained by using FLMs are statistically significantly better than those by using only the baseline 3-gram model. However, the different FLMs do not lead to statistically significantly different results.

\section{Conclusion and Future Work}
\label{sec:conclusion}
The factored language models outperform a traditional 3-gram language model both in terms of perplexity and mixed error rate on the SEAME Code-Switching corpus. The combination of the features open class words, Brown word clusters and POS tags achieves the best perplexity results on the development and evaluation sets and the best mixed error rate results on the evaluation set. Brown word clusters alone lead to a similar performance as POS tags alone. Their advantage is that they do not rely on an expensive tagging process with unknown accuracy. On the development data, the combination of Brown word clusters, POS tags and clusters of open class word embeddings leads to the best mixed error rate results. Most of these improvements are also statistically significant.\\
Although the method and features are evaluated only on a Mandarin-English Code-Switching corpus in this paper, the methodology is language pair independent. Hence, it can be applied to corpora with different languages, too. Especially the Brown word clusters and open class word clusters do not require knowledge about the language of a certain word.\\
Possible future work is the integration of machine translation in order to create monolingual corpora based on the bilingual text and extract additional features from them.

\section*{Acknowledgment}
The authors would like to thank Dr Li Haizhou to allow us to use the SEAME corpus for this research work.

\ifCLASSOPTIONcaptionsoff
  \newpage
\fi



\bibliographystyle{IEEEtran}
\bibliography{thesis}
%



%
\vspace{-1.2cm}
\begin{IEEEbiographynophoto}{Heike Adel}
is a PhD student at the Center for Information and Language Processing, University of Munich in Germany. She studied Computer Science at Karlsruhe Institute of Technology (KIT) in Karlsruhe, Germany and received her Bachelor degree in 2011 and her Master degree in 2014. Since April 2014, she works as a research and teaching assistant at University of Munich under the supervision of Prof. Hinrich Schuetze. The main focus of her work is natural language processing using deep learning techniques.
\end{IEEEbiographynophoto}
\vspace{-1.2cm}
\begin{IEEEbiographynophoto}{Ngoc Thang Vu}
received his PhD in Computer Science from the Karlsruhe Institute of Technology, Germany in 2014. He joined Nuance Communications as a senior research scientist in 2014. Currently he is also a visiting professor in Computational Linguistics  at the University of Munich (LMU). His research interests are multilingual speech recognition for low resource languages and accents, and natural language processing.

\end{IEEEbiographynophoto}
\vspace{-1.2cm}
\begin{IEEEbiographynophoto}{Katrin Kirchhoff}
received her PhD in Computer Science from the University of Bielefeld, Germany, in 1999.  She is currently a
Research Associate Professor in Electrical Engineering at the
University of Washington. Her research focuses on speech recognition,
natural language processing, machine translation, and human-computer
interfaces. She has authored or co-authored over 100 publications on
speech and language processing. From From 2009-2011 she was a Member
of the IEEE Speech Technical Committee.  She currently serves on the
editorial boards of Speech Communication and Computer, Speech and
Language. 
\end{IEEEbiographynophoto}
\vspace{-1.2cm}
\begin{IEEEbiographynophoto}{Dominic Telaar}
is a research assistant at the Cognitive Systems Lab, at the Institute
of Anthropomatics and Robotics at the Karlsruhe Institute of Technology (KIT) in Germany.
He received his Diploma degree in Computer Science at KIT in 2010.
Afterwards, he has worked as a research and teaching assistant at KIT. The
main focus of his work is the development of techniques for the BioKIT recognition
toolkit.
\end{IEEEbiographynophoto}
\vspace{-1.2cm}
\begin{IEEEbiographynophoto}{Tanja Schultz}
received her Ph.D. and Masters in Computer Science from University of Karlsruhe, Germany in 2000 and 1995 respectively. She joined Carnegie Mellon University in 2000 and became a Research Professor at the Language Technologies Institute. Since 2007 she is a Full Professor at the Department of Informatics of the Karlsruhe Institute of Technology (KIT) in Germany. Her research activities focus on human-machine interfaces with a particular area of expertise in rapid adaptation of speech processing systems to new domains and languages. She has published more than 250 articles in books, journals, and proceedings, and has received several awards and prizes for her work. She is a member of the Society of Computer Science (GI) for more than 20 years, of the IEEE Computer Society, and the International Speech Communication Association (ISCA) where she was elected as the president in 2013.
\end{IEEEbiographynophoto}

\end{document}